# An information theoretic limit to data amplification

S. J. Watts[*], L. Crow

Department of Physics and Astronomy, The University of Manchester, Oxford Road, M13 9PL, Manchester, UK

*Correspondence to: Stephen.Watts@manchester.ac.uk

**Abstract:** In recent years generative artificial intelligence has been used to create data to support science analysis. For example, Generative Adversarial Networks (GANs) have been trained using Monte Carlo simulated input and then used to generate data for the same problem. This has the advantage that a GAN creates data in a significantly reduced computing time. $N$ training events for a GAN can result in $NG$ generated events with the gain factor $G$ being more than one. This appears to violate the principle that one cannot get information for free. This is not the only way to amplify data so this process will be referred to as *data amplification* which is studied using information theoretic concepts. It is shown that a gain of greater than one is possible whilst keeping the information content of the data unchanged. This leads to a mathematical bound, $2\log(\text{Generated Events}) \geq 3\log(\text{Training Events})$, which only depends on the number of generated and training events. This study determines conditions on both the underlying and reconstructed probability distributions to ensure this bound. In particular, the resolution of variables in amplified data is not improved by the process but the increase in sample size can still improve statistical significance. The bound is confirmed using computer simulation and analysis of GAN generated data from the literature.

1. Introduction and Motivation

Monte Carlo simulation has become a vital in many science areas. It is very computer intensive, for example, the ATLAS experiment at the CERN Large Hadron Collider (LHC) spent 50% of its CPU resources on Geant4 simulations in Run 2, [1]. Upgrades to the LHC and its experiments will result in considerably more data and thus the need for even more simulation. The required computer resources do not exist thus much effort has been devoted to optimising Geant4 and developing fast simulation alternatives for less critical applications. Moreover, there is an environmental cost to this amount of computing. A radical solution is the use of AI in the form of Generative Adversarial Networks. These are trained with $N$ events from a full Geant4 Monte Carlo simulation and then used to generate $GN$ events, where $G$ is the gain. The trained GAN can generate events a few orders of magnitude faster on the same CPU/GPU compared to the full Monte Carlo simulation. GANs are being used across a wide range of disciplines from medical applications to remote sensing, [2]. Initially they were applied to images. However, in medical physics dose calculations, particle physics and many other areas, it is vital that " the distribution (probability density function) is correct", [3]. This idea is being applied in medical physics [3, 4, 5], astrophysics [6] and particle physics [7, 8, 9, 10]. In some applications the GAN model is used to generate a compact dataset that describes the system and this is used to quickly generate events as required [5]. The phrase *simulation* will mean the use of a Monte Carlo method. *Generation* will mean that a trained AI algorithm has generated the data.

Taking particle physics as an example, [7] applied the technique to the generation of electromagnetic showers – considered as jet images - and found a speedup of ~14 compared to a PYTHIA Monte Carlo simulation when using a CPU. The GAN method can use a GPU which improves the speedup by another factor 15. This suggests that one might be able to greatly speed up simulation if the GAN model generates more data than was required to train the algorithm. The application to calorimetry is discussed in [8] which used the Geant4 Monte Carlo for simulation. This gave a speedup factor of ~800 using CPUs with an extra ~ 170 using GPUs. A study of the so called "*GANplification*" process to toy problems is given in [9]. In [10] the method is applied again to calorimetry and compared to Geant4 simulation and called "*Calomplification*". This paper will use the generic term, "*data amplification*" for training any algorithm with a dataset – real or simulated – such that it can generate a new dataset with $G \geq 1$. This has the potential to substantially reduce computing time. However, this appears to violate the principle that one cannot generate information out of nothing, which was much discussed after talks by Ramon Winterhalder and Lukas Heinrich at a meeting of particle physicists, astrophysicists and statisticians at the PhyStat Workshop, held in London in September 2024, [11].

This paper will show that data amplification works and the information content is not changed. The penalty is that it trades the resolution in each variable for increased statistics, i.e. more events. This paper will place a bound on the maximum gain, describe the conditions required of the underlying pdfs to ensure this bound, and illustrate how to validate generated data.

2. Background theory

This section derives and discusses the bound on the gain mentioned in the abstract and tested in Sections 3 and 4. The information associated with data is understood by studying the Shannon entropy of a histogram for a continuous variable, which is described in Sections 2.1 and 2.2. This forms the foundation for the derivation in Section 2.3 of the bound on data amplification.

2.1 Histogram algorithm and its Shannon entropy

An algorithm for finding the bin width of a continuous variable using entropy is discussed in [12, 13, 14]. A full description with worked examples on many types of data is given in [13]. It is based on the Shannon entropy and differential entropy. The first applies to binned data. The second to the underlying probability density function (pdf), $p(x)$.

The differential entropy, $h$, of continuous data is,

$$h = -\int_S p(x)\log(p(x))dx \tag{2.1}$$

$h$ can be evaluated from data using a kNN non-parametric estimator, [15]. If the data is histogrammed then the binned (discrete) Shannon Entropy is,

$$H = -\sum_{i=1}^{i=N_{Bin}} p_i \log(p_i) \tag{2.2}$$

$H_B$ is an estimate of $H$ from the data, with the probabilities, $p_i$, being estimated using $n_i$ the actual the number of entries in bin $i$ and $N_{Bin}$ being the total number of bins. For a uniform distribution, define a parameter $M$ which relates the number of bins, $N^{1/M}$, in terms of the total number of entries or events, $N$. Clearly $M \geq 1$ and as will be shown, this controls the $N$ dependence of the bin width.

For a uniform distribution, the average number of entries per bin, $\mu_H = \frac{N}{N^{1/M}} = N^{(1-1/M)}$ (2.3)

The entropy for this is simply, $H = \frac{1}{M}\log N$ (2.4)

Remarkably Eq. (2.4) is true for all pdfs [13]. $N_{Bin} \geq N^{1/M}$ with equality only for the uniform distribution. A new and simple derivation of Eq. (2.4) is given in Appendix A. For a histogram with fixed bin width, $\Delta$, one can derive the relationship between $h$ and $H$,

$$H = h - \log(\Delta) \tag{2.5}$$

which is also derived in Appendix A. Combining Eq. (2.4) and Eq. (2.5) gives,

$$\Delta = \exp(h - H) = \frac{\exp(h)}{N^{1/M}} \tag{2.6}$$

This is an exact solution for the bin width to use for a continuous variable. The only issue is the choice of $M$. $M$ is a "Goldilocks" statistic as it must be in a certain range as explained in Section 2.2. For a uniform pdf, $M = 2$, as this pdf has no shape. For all other pdfs, $2 \leq M \leq 3$, to avoid over-binning or under-binning. There are two ways to assure this.

1) Use Eq. (2.6) and set the bin width appropriately.
2) Bin the data using another algorithm and check the value of $M$ by using the estimate of $H = H_B$

$M$ can be estimated as either $M_B = \frac{\log(N)}{H_B}$ (2.7)

or $\qquad H \simeq H_X(bits) = \log_2(\frac{N}{N_{Max}}) + 1 \quad M_X = \frac{\log_2(N)}{H_X}$ (2.8)

The second result is derived in [13]. This gives a good estimate of the histogram entropy in bits and requires just the maximum bin entry and total number of entries in the histogram. The $+1$ in Eq. (2.8) is zero for a uniform pdf.

Eq. (2.4) shows that continuous data has a well-defined Shannon entropy, or information content, because the data is finite. It is at most $(1/2)\log_2(N)$, which means that collecting four times more data increases the entropy by one bit for a single variable. In this case, errors usually improve by a factor $\sqrt{4} = 2$ which is also equivalent to a bit.

**2.2 Choice of the "Goldilocks" parameter $M$**

$M$ changes the bin width which gets larger as $M$ increases. Excess Poisson noise (over-binning) is removed once $M \geq 2$ and is pdf independent, [13]. As the bin width gets larger, the shape of the pdf is not correctly reconstructed (under-binning). This happens for $M > 3$. This is understood due to the work of Shimazaki and Shinomoto, [16]. They derive a *cost function* which is written as,

$$C_{SS\Delta} = \tfrac{1}{N\Delta} - \varphi(0) - \tfrac{1}{3}\varphi'(0)\Delta - \tfrac{1}{12}\varphi''(0)\Delta^2 + O(\Delta^3) \tag{2.9}$$

$\varphi(\tau)$ is the autocorrelation function of the pdf. It is a measure of how each point on the distribution is related to points nearby. Two approximate solutions are found. One for the case when $\varphi'(0) \neq 0$ and $\varphi''(0) = 0$. This corresponds to $\varphi(\tau)$ having a cusp at $\tau = 0$. This gives an $M = 2$ solution ($\Delta \propto N^{-1/2}$) and the key example of this is the uniform distribution. Well behaved distributions usually have $\varphi'(0) = 0$ and this gives an $M = 3$ solution ($\Delta \propto N^{-1/3}$). This can be shown to be the same as Scott's result, ref. [13, 17]. In conclusion, apart from a uniform distribution, the minimum of the cost function is between $M = 2$ and $M = 3$. A cost function, $C_B$ based on Eq. (2.9) is derived in [16], which can be estimated from data,

$$C_B = \tfrac{2\mu_B - \sigma_B^2}{\Delta^2} \tag{2.10}$$

$\mu_B$ is the mean number of events per bin.
$\sigma_B^2$ is the variance on the number of events per bin

This can be estimated as a function of $M$ by varying the bin width. $C_B$ is maximum at $M = 1$ with $M = 2$ being the point at which excess Poisson fluctuations are gone. Consequently, it is useful to normalise $C_B$ by referencing it to $M = 2$ and scaling it to one between $M = 1$ and $M = 2$. This is the normalised cost function shown in Fig. 1. The cost function is almost flat between $M = 2$ and $M = 3$ for the standard normal and Moyal distributions, so a choice of $M$ in this region will have little effect on the knowledge of the pdf. The minimum is reached for the uniform distribution at $M = 2$ and remains unchanged thereafter since it has no shape. The curve illustrates the range of $M$ for which the pdf is best approximated by the bin width. Applying the normalisation conditions to Eq. (2.9), substituting $\Delta$ using Eq. (2.6) and neglecting small terms, one arrives at,

$$C_{NR} = \tfrac{1}{\mu_H} - N^{-1/2} - \tfrac{1}{12}\varphi''(0)\exp(3h)\left(\tfrac{1}{N^{2/M}} - \tfrac{1}{N}\right) \tag{2.11}$$

This assumes that $\varphi'(0) = 0$ which is true for most well-behaved pdfs. The first term depends only on the pdf independent parameter $\mu_H$ defined in Eq. (2.3). The cost falls rapidly with $M$ as the excess Poisson fluctuations are removed. The cost functions in Fig. 1 were fitted with,

$$C_{NR} = \tfrac{1}{\mu_H} - N^{-1/2} + A\left(\tfrac{1}{N^{2/M}} - \tfrac{1}{N}\right) \tag{2.12}$$

which just depends on the parameter $A$ with a theoretical value,

$$A = -\tfrac{1}{12}\varphi''(0)\exp(3h) \tag{2.13}$$

$A$ is positive as $\varphi''(0)$ is negative for sensible pdfs. The fits are excellent. To two decimal places, the expected values of $A$ are 0, 0.83, 1.48 and 11.80 for the uniform, standard normal, Moyal, and standard log-normal distributions respectively. Fitted values were 0, 1.31, 2.76, and 12.06 respectively. $A$ is independent of the standard deviation for the normal distribution since the dependence due to the standard deviation term in $\varphi''(0)$ and $\exp(3h)$ cancel out. The Moyal distribution, ref. [13], has no scale or location parameters.

Fig. 1 shows that the log-normal is a distribution that needs $M$ closer to 2. Not because it has a cusp in $\varphi(\tau)$ but because $\varphi''(0)$ is large due to its significant kurtosis. This is a good example of a distribution that is difficult to histogram with a fixed bin width due to its large kurtosis. The solution to this problem is described in [13] by changing the variable using Box-Cox [18] or Yeo-Johnson [19] to transform the pdf to a normal distribution. These useful transformations find application in Section 3.

**2.3 Limit to data amplification**

Training data - either Monte Carlo or real - is created for $N$ events. Initially consider the problem in one dimension. The pdf is estimated using a histogram with fixed bin width, $\Delta$. Eq. (2.6) is used to calculate this bin width and the resulting reference histogram, with $M = M_0$ has,

$$\Delta = \exp(h)N^{-(1/M_0)}$$

This estimated pdf is used to generate new events by some algorithm. A simple method will be described in the next section. The number of generated events is $GN$ with the gain, $G \geq 1$. The pdf of the generated events is estimated with a histogram, with a bin width that is identical to the reference.

$$\Delta = \exp(h)(GN)^{-1/M}$$

Note that the pdf of the generated histogram must be the same as the reference thus both have the same value of $h$ thus,

$$\exp(h)(GN)^{-1/M} = \Delta = \exp(h)N^{-(1/M_0)} \text{ and } (GN)^{-1/M} = N^{-(1/M_0)}$$

Thus $GN = N^{(M/M_0)}$ and $G = N^{\left(\frac{M}{M_0}-1\right)}$ \hfill (2.14)

This can be re-written as, $\log(GN) = \tfrac{M}{M_0}\log(N)$ , from which,

$$\tfrac{1}{M}\log(GN) = \tfrac{1}{M_0}\log(N) \tag{2.15}$$

This means that the Shannon Entropy ( or information) of the generated histogram is the same as the reference histogram. However, the estimates of the pdf will only be acceptable if $M$ and $M_0$ are at the minimum of the cost curve. To a good approximation one can achieve this, due to the flatness of this curve between $M = 2$ and $M = 3$.

Taking $M_0 = 2$ as the reference point at which excess Poisson fluctuations no longer affect the cost then,

$$M = M_0 \frac{\log(GN)}{\log(N)} = 2\frac{\log(GN)}{\log(N)}$$

This value of $M$ will be defined as the effective value, $M_{eff} \equiv 2\frac{\log(GN)}{\log(N)}$ \hfill (2.16)

This parameter can be estimated for any data amplification algorithm as it only depends on the number of reference (real data or training data) events and the number of generated events. This result contains no parameters associated with the amplification algorithm and should apply in general.

For any amplification algorithm (e.g. GAN), $M_{eff} \equiv 2\frac{\log(\text{Generated Events})}{\log(\text{Training Events})}$ \hfill (2.17)

The base of the log is irrelevant. For the generated events, one must restrict $M_{eff} \leq 3$, if the cost function of the single variable is to remain in the minimum range. This means that one can achieve $G > 1$ and preserve the pdf provided,

$$2\frac{\log(\text{Generated Events})}{\log(\text{Training Events})} \equiv M_{eff} \leq 3 \hspace{2cm} (2.18)$$

This will be referred to as the *amplification bound.* Amplification is possible because the Shannon entropy of a finite number of samples from a continuous pdf is dependent on both $M$ and $N$. To extract reliable results using this pdf requires an unbiased and low cost reconstruction of its shape and also *good statistics* or a sufficient number of events. One can trade-off cost controlled by $M$, with events, $N$ and the information content is unchanged. This will achieve a good parameterisation of the shape of the pdf because in most cases it is smooth and well behaved. These factors will be explored using a simple amplification algorithm in Section 3.

Clearly, one cannot amplify amplified data because the intrinsic resolution of each variable, $\Delta \sim \frac{\exp(h)}{\sqrt{N}}$, from the training data stays with the amplified data. This is the penalty paid for this process. See Section 3 for more detail.

Finally, this result applies to multivariate data. The input training data consists of n-tuples of $N$ events. There are the same number of entries for each variable. The amplification algorithm applies the same gain to the $N$ events and thus all variables. From Eq. (2.14), this is only possible if the same factor $M/M_0$ applies to all variables.

## 3. Comparison with a simple data amplification algorithm

This algorithm follows naturally from the theory section. The algorithm is simple and although it has limitations it illustrates the key ideas.

The algorithm is applied to a standard normal distribution, $N(0,1)$, and a standard log-normal, $LogN(0,1)$. It is described in pseudo code below. This algorithm is shown diagrammatically in Fig. 2. All analyses were performed using the R Statistical Software v4.4.1, [20], within the RStudio framework, [21]. Two problems are clear. First, the range of the generated data is the same as the original version. Second, although the datasets are random copies and not identical copies, the number of events in each histogram bin is scaled up by a factor $G$ and thus the bin contents error is not scaled correctly. The first problem was minimised by starting with a reasonable value of $N = 2000$. The second problem is discussed later. Next, $GN$ events are simulated directly from a random generator according to the selected choice of pdf. This is the *reference dataset*.

The *generated data* was compared with the *reference dataset* by evaluating the Kullback-Leibler divergence (KLD) between these two sets of data using a kNN estimator which is built into the FNN package in R, [22]. KLD is zero for identical distributions.

---

**Algorithm** Generate a training dataset with a specific pdf and create random copies of this training set and concatenate to create a generated dataset.

---

**Input**: $N$ the number of points in the dataset to be copied. This is called the *training data*.
  Choice of pdf – in this paper, $N(0,1)$ or $\text{LogN}(0,1)$.
**Output**: $GN$ datapoints generated from the training data with integer gain $G = 1, 2, 3\ldots\ldots$
  This is the *generated training set*.
**Procedure** $\text{GenCopy}(N, G)$

1. Simulate $N$ random points with a specific pdf (Normal or Log-Normal).
2. Bin the values using Eq. (2.6) with $M = 2$.
3. Generate a randomised copy of this data by creating $x = x_{Start} + (i + rnd)\Delta$ with $x_{Start}$ as the starting point and $\Delta$ as the bin width. $rnd$ is a uniform random number between zero and one. The index, $i$, runs from 1 to $N_{bin}$ and is the bin number for the original data point.
4. Make $G$ copies and concatenate them into a single dataset.

**End Procedure**

---

It is important to note that the KLD is calculated from the data values in $x$ and not from a histogram. Writing $g(x)$ for the pdf of the generated data, KLD can be written formally as $D(G \| P)$. This divergence is not always symmetric to an interchange of $G$ and $P$. However, for the distributions evaluated it is symmetric so just one value is shown. The same kNN estimator for KLD was used and coded independently by the authors in [12,13,14]. Fig. 3a) shows the divergence as a function of $M_{eff}$ for the normal distribution. All gain values up to $M_{eff} = 3$ are calculated. After this, a few values around $M_{eff} = 3.1$ to $4.0$ in steps of $0.1$ are calculated. This is to reduce the computing time. $G = 1, 7, 45, 299$ and $2000$ at $M_{eff} = 2, 2.5, 3, 3.5$ and $4$ respectively. The KLD is small and the errors are large for $2 \leq M_{eff} \leq 3$. To get a better estimate, a method evaluated in detail in [14] was employed. For each dataset with $GN$ points, several copies were made using the algorithm above for both the amplified and reference data. KLD was then evaluated for an average over $16$ copies or iterations, $N_I$. The averaged value is also shown in Fig. 3a) and the error is considerably reduced. The kNN estimator is unbiased and consistent and so is this averaged estimate, see [14] for details. The variance model for the differential entropy, $h$, [15] and mutual information (MI), [14] are,

$$Var(h) = \frac{1}{N}\left[Var(\log(p(x))) + \Psi_1(k)\right] \qquad (2.19)$$

$\Psi_1(k)$ is the trigamma function which for $kNN = 1$ and $4$ is 1.645 and 0.284 respectively.

$$Var(MI) = \frac{3 \times Var(h)}{NN_I} + \frac{1+\rho^2}{2N} \qquad (2.20)$$

The factor three is because the mutual information comprises three calculations of $h$. $\rho$ is the correlation between the two variables. KLD is the mutual information between the joint pdf and the product of the marginal distributions. For this reason, $\rho = 0$ and the final variance model for a normal distribution is,

$$Var(KLD) = \frac{1}{N}\left[\frac{6.44}{N_I} + \frac{1}{2}\right] \quad (\text{nats})^2 \qquad (2.21)$$

Using Eq. (2.21) the error for each value of $M_{eff}$ is estimated for just one iteration or 16 iterations. This is shown in bits in Fig. 3b). The second term in Eq. (2.20) and Eq. (2.21) is the residual error due to the approximation to the pdf from the finite bin width and cannot be eliminated by iteration, labelled in Fig. 3b) as the error base. The fluctuations in KLD in Fig. 3 a) are consistent with the error calculation in Fig. 3 b). This averaging procedure is very effective and is similar to the data amplification process. Both eventually reach an error floor due to the remaining uncertainty in the pdf due to the finite bin width.

Fig. 3a) is the first demonstration of the *amplification bound* in Eq. (2.18). The small but non-zero KLD from $2 \leq M_{eff} \leq 3$ is due to the failure to model the tail correctly. For $M_{eff} > 3$, the KLD grows linearly with $M_{eff}$. This is consistent with the cost function growing beyond this point as shown in Fig. 1.

The same procedure was followed for $LogN(0,1)$. The result is shown in Figure 4a). The KLD for this distribution as a function of $M_{eff}$ is shown for both $kNN = 1$ and $kNN = 4$. The generated data is significantly different to the reference, when compared to the $N(0,1)$ case. The $kNN = 4$ KLD is smaller because on a longer distance scale the pdf is better reconstructed. The fluctuations on the $kNN = 4$ points are also smaller than the $kNN = 1$ results. This is due to the smaller trigamma function term in Eq. (2.19). The KLD shows that for $M_{eff} > 2.5$ the pdf is less well reconstructed. This agrees with the cost function analysis in Section 2.2 and Fig. 1. A key property of MI and KLD is that when the pdfs undergo a smooth and uniquely invertible transformation they are unchanged, [23]. Ref. [24] has a formal proof. This means that the KLD for $LogN(0,1)$ in Fig. 4 should be similar to Fig. 3, but this clearly fails due to the different cost function for the histogram estimate of the pdf, shown in Fig. 1. However, one can transform the standard log-normal distribution to a normal distribution. The transformation maps the $LogN(0,1)$ to $N(0,1)$ by setting $x' = \log(x)$. The transformed distribution is then amplified. It is transformed back to $LogN(0,1)$ by setting $x'' = \exp(x')$. This is then compared to a reference $LogN(0,1)$ distribution. The transform is a Box-Cox transformation, [18]. The mapped result is shown by the lower curve in Fig. 4a) which is compatible with the normal distribution calculation in Fig. 3a). It is now possible to obtain a very small KLD value up to $M_{eff} = 3$.

To relate these numbers back to the gain, Fig. 4b) replots the data in Fig. 4a) versus gain rather than $M_{eff}$. The vertical lines correspond to $N^{1/4}$ and $N^{1/2}$ or $M_{eff} = 2.5$ and $M_{eff} = 3.0$ respectively in Eq. (2.14) with $M_0 = 2$. The clear difference between the direct amplification of $LogN(0,1)$ and a mapped $\log N(0,1) \to N(0,1) \to \log N(0,1)$ method is very clear. This is an important result since it demonstrates that it should be possible to amplify even the most difficult pdf *should such a mapping be possible.*

Next a test was made to confirm that amplified data should not be used as training data. This can be understood in terms of the following mathematics. Training data with $N$ events is pre-amplified to $GN$ events with $M = M_P$. This is further amplified by $G'$ to $G'GN$ events with $M = M_A$. These processes can be written,

$$\frac{1}{2}\log(N) \Rightarrow \frac{1}{M_P}\log(GN) \Rightarrow \frac{1}{M_A}\log(G'GN) \quad \text{and for the last amplification,} \quad M_{eff} = 2\frac{\log(G'GN)}{\log(GN)} \qquad (2.22)$$

The arrows show the direction in which these processes are followed. These arrows are also equal signs and, $2 \leq M_P \leq M_A \leq M_{eff}$. If $G = 1$ then the bound is $M_{eff} = 3$. Simple algebra shows that,

$$M_{eff} = 2M_A / M_P \qquad (2.23)$$

If there is no pre-amplification, then $M_{eff} = M_A$. This situation has already been modelled. The bound on pre-amplified data is $M_{eff} = 6/M_P$. $M_P \geq 2$ which means that for data that has been pre-amplified, $M_{eff} < 3$. An increase in the KLD before $M_{eff} = 3$ implies either that the underlying pdf is problematic – e.g. log-normal – or that the data may have been pre-amplified. The processes described by Eq. (2.22) were modelled for a standard normal distribution, $N(0,1)$, with $N = 500$, $G = 4$ and the resulting $GN = 2000$ data was amplified. The result for KLD as a function of $M_{eff}$ for the pre-amplified and then amplified data is compared to a fully simulated reference in Fig. 5. The pre-amplified data is trained with just 500 points and the tail is less well described. This leads to a region between $2 \leq M_{eff} \leq 2.5$ in which the KLD is larger than the $N = 2000$ fully trained data. The twice amplified data shows increasing KLD beyond $M_{eff} \sim 2.5$, consistent with expectation, since, $M_P = 2.446$ and using Eq. (2.23) the bound is $M_{eff} = 2.453$. $N = 2000$ with $G = 1$ results are taken from Fig. 3a) for comparison in which the maximum $M_{eff}$ before the onset of an increase in KLD is around 3.0. The overall gain is, $GG' = 4 \times 5.59 = 22.4 = \sqrt{500}$. This confirms that amplified data remembers the initial number of training events, and thus the resolution is set by this sample size, $\Delta \approx \exp(h)/\sqrt{N}$. It can only be amplified further if its initial $M_P \leq 3.0$ although this would not be recommended.

The final check was to perform a parametric fit to the amplified distribution to quantify how well it agreed with expectation. This is shown for $LogN(0,1)$ with a training input of $N = 2000$ and $G = 45$. This corresponds to $M_{eff} = 3$. Fig. 6 compares a fully simulated $N = 90000$ reference distribution (filled circle) with an amplified $LogN(0,1)$ (filled square) and amplified mapped $LogN(0,1)$ (half filled square). There are clear errors in the right hand tail for the amplified $LogN(0,1)$ which are improved using the mapping method. However, there is a more subtle fault in the amplified data. This is revealed by the $\chi^2/DF$ of the fit. The fit to the data only allowed the overall amplitude to be a free parameter. The location and scale parameters were for $LogN(0,1)$. The Baker and Cousins Poisson chi-squared, [25], was used as this is robust to zero entries. Their chi-squared formula is,

$$\chi^2_{\lambda,p} = 2\sum_i \left[ y_i - n_i + n_i \log(n_i/y_i) \right] \qquad (2.24)$$

$y_i$ is the predicted $y$ value in bin $i$ with entry $n_i$. The results of the fits are shown in Table 1.

**Table 1** Results of curve fits to the log-normal distribution in Figure 6. $M_{eff} = 3$

| Distribution | Gain, G | N | Amplitude | $\chi^2_{\lambda,p} / DF$ | Range |
|---|---|---|---|---|---|
| $LogN(0,1)$ | Original, 1 | 90000 | 3319.5 +/- 3.7 | 447.0/399 = 1.12 | $0 \leq x \leq 40$ |
| $LogN(0,1)$ | 45 | 2000 | 3290.3 +/- 13.5 | 5730.5/399 = 14.36 | $0 \leq x \leq 40$ |
| $LogN(0,1)$ Mapped | 45 | 2000 | 3255 +/- 15.9 | 2168/399 = 5.43<br>2016/236 = 8.6 | $0 \leq x \leq 40$<br>$0 \leq x \leq 20$ |

The amplified $LogN(0,1)$ has a poor $\chi^2_{\lambda,p}/DF$ because $M_{eff}=3$ which as Fig. 1 shows has a large cost. The amplified and mapped $LogN(0,1)$ is much better, however, the $\chi^2_{\lambda,p}/DF$ is still too large. The fully simulated $LogN(0,1)$ has an excellent $\chi^2_{\lambda,p}/DF$. As noted earlier, the data amplification is kept within the bounds of each bin. This means that the error on each bin is $G\sqrt{N}$ rather than $\sqrt{GN}$. It is a factor $\sqrt{G}$ larger than it should be or $\sqrt{45}=6.7$. This is confirmed by the result in Table 1. The mapped distribution is estimated over two ranges as it has a smaller tail. The conclusion is unchanged. Despite its deficiencies, this simple algorithm has displayed the key results expected from the analysis in Section 2.

**4. Comparison with GAN results**

The amplification model used in Section 3 was kept simple to demonstrate the theory in Section 2 that resulted in the amplification bound, Eq. (2.18). Some deficiencies can be fixed by using the histogram estimate to create a cumulative density function and generating data using this. A kernel density estimate approach would improve the modelling of the pdf tails. However, the key reason to start with this simple algorithm is that it has a well-defined initial value for $M$ since the bin width is chosen using Eq. (2.6). The bin width for $M=2$ is the smallest value that the pdf approximation can probe. Below that excess Poisson noise leads to a poor pdf description. It is often assumed that the smallest value to which a variable can be probed is set by the experimental resolution. However, this cannot be reached unless enough data has been recorded. Nowadays, AI deep learning exists with excellent methods that learn and parametrize the functional form of distributions. They are adept at dealing with non-linear variables. This route to an amplification algorithm seems most effective. A reasonable assumption is that even these algorithms cannot probe at variable sizes below the limit set by Eq. (2.6). Thus one can argue that after a GAN is trained with $N$ events, that $M_0=2$ and the smallest size probed is set by Eq. (2.6). As argued in Section 2, Eq. (2.18) should apply to any algorithm. To validate this argument, work on GAN amplification ( " GANplification") has been reviewed to test Eq. (2.18).

First, the use of GAN models to generate calorimeter data in particle physics, [10]. The authors compare different distributions associated with calorimetry using a GAN generator trained with $N$ events. This is compared to a full Monte Carlo simulation using the Geant4 package. Distributions are compared using quantiles with equal probability. The Jensen-Shannon divergence was used to compare the generated and reference distributions. This is a symmetrised version of the Kullback-Leibler divergence. They summarise their results in the conclusion as, " For a training sample of 1k showers we generate up to 1000k showers from the network and find a comparable performance of up to 50k Geant4 showers for the kinematic distributions and their correlations ". The limit of 50k is deduced from their figs. 6 and 7. Applying Eq. (2.18)

$$M_{eff}=2\left(\frac{\log(50000)}{\log(1000)}\right)=3.13$$ which is at the upper level of the bound derived in Section 2.

Second, [9], studied the use of GAN with three toy models with $N=100$ and $500$. They studied the quantile error as a function of the number of generated GAN events. The models were a $1D$ camel back, $2D$ Gaussian ring and $5D$ Gaussian spherical shell. The error drops with increasing number of generated events and reaches a floor once the GAN model is tested at its smallest detail. This is similar behaviour to the formula for the iterated error on the Kullback-Leibler divergence estimate in Eq. (2.21) where the copies reduce the error but eventually the limit due to variable resolution is reached. Some selected data is re-plotted in Fig. 7a) from Ref. [9]. This is then plotted as a function of $M_{eff}$ in Fig. 7b). The floor is found at $M_{eff}=3$ for all the models. This is true for all data plotted in Fig. 2, Fig. 4, and Fig. 6 of [10] which are for single variables; 2D $(radius,\phi)$, 5D $(radius,\phi_1,\phi_2,\phi_3,\phi_4)$. A factor ten amplification factor (or $M_{eff}=3$) is assigned by [10] to their Fig. 2. This shows the lack of any dimensionality dependence predicted by Eq. (2.18).

## 5. Summary and Conclusions

Three key equations will be repeated for clarity. This paper shows that data amplification is possible due to a bound on the maximum gain which is independent of the dimensionality of the data,

$$2 \frac{\log(\text{Generated Events})}{\log(\text{Training Events})} \equiv M_{eff} \leq 3 \tag{2.18}$$

Clearly, the gain can be greater than one. This was derived using an information theoretic argument. The Shannon entropy of a variable with $N$ entries is,

$$H = \tfrac{1}{M} \log N \tag{2.4}$$

A value of $2 \leq M \leq 3$ ensures that knowledge of the pdf will have minimum cost. $M$ also controls the bin width or smallest resolution at which the pdf is known. One can trade $M$ for $N$ by keeping the resolution fixed. As $M \uparrow$ then $N \uparrow$ to keep the entropy, $H$, constant. The smallest variable resolution is set by $M = 2$ or explicitly,

$$\Delta \sim \frac{\exp(h)}{\sqrt{N}} \text{ which for a normal distribution is, } \Delta \sim \frac{4 \times \text{Standard Deviation}}{\sqrt{N}} \tag{2.23}$$

The method works because most pdfs are smooth and well behaved. This statement can be checked by estimating the autocorrelation of the pdf or studying the cost as a function of $M$. Even troublesome pdfs can be tamed by transforming them to a normal distribution using a Box-Cox or Yeo-Johnson transformation. The algorithm also needs to correctly model the tails of the pdf. It is assumed that AI algorithms such as GAN's will find such a solution as deep neural networks are very good at learning non-linear functions. Eq. (2.18) was verified with a simple data amplification algorithm using the Kullback-Leibler divergence to quantify how close amplified data was to a fully simulated pdf. A fit to the pdf is important to check that the errors behave as expected. Eq. (2.18) was found to apply to results in previous papers, [9, 10], which used GAN amplification.

The penalty paid – the resolution on the variable, Eq. (2.23) - for this amplification should not matter in most applications. In many cases, statistical significance depends mainly on the number of events. For example, the method to reduce the error on the Kullback-Leilber divergence estimate, Eq. (2.21), benefited from averaging over many copies of the original data and reached a floor due to the resolution set by the histogram estimate of the pdf. Without extra copies of the data this limit would not have been achieved.

One cannot amplify the generated data if one has already reached the bound. The intrinsic resolution is built into the amplified data from the initial training values, so it is inadvisable to repeat this process.

Information theory proved vital to understand this problem. Shannon's work led to advances in data compression which are especially useful for images and for which GAN's have been especially effective at generating. Expanding compressed JPEG data to a full size image is similar to data amplification. JPEG is a lossy algorithm but this can be tolerated as the final image is usually sufficient for further analysis.

### Acknowledgements

We thank John Allison and Cinzia Da Vià for useful discussions. LC thanks UKRI (STFC) and The University of Manchester for research student funding. SJW thanks the Leverhulme Trust for their support with an Emeritus Fellowship.

# FIGURES

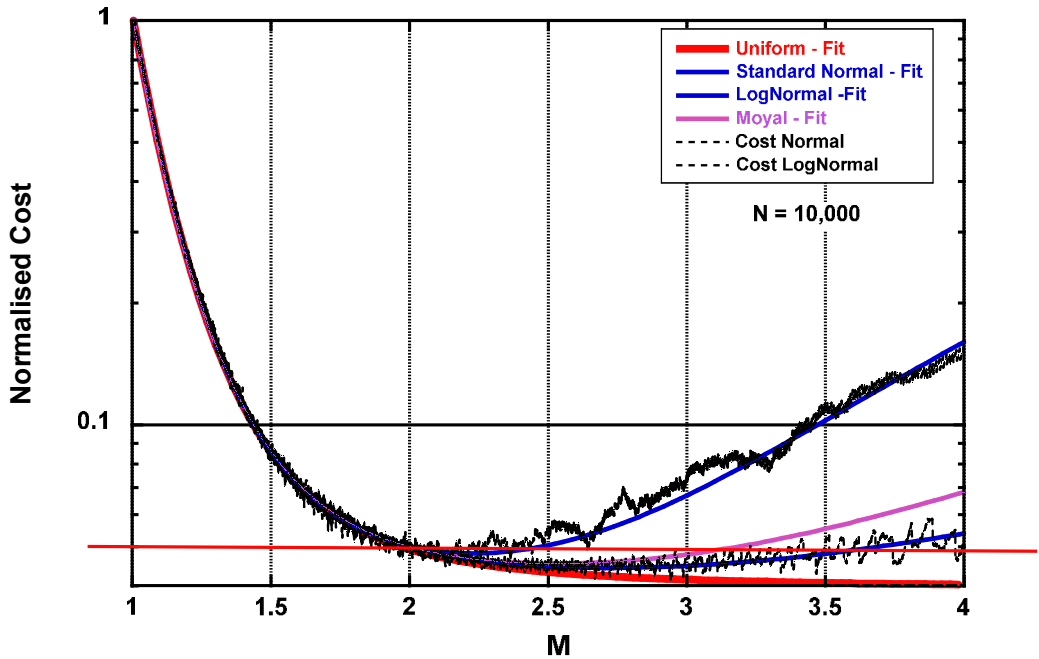

**Figure 1.** Normalised cost as a function of the parameter $M$ ( see text). The fits to simulated data for 10,000 events are shown for uniform, standard normal, Moyal and standard log-normal distributions. The curves are in this order of increasing cost on the right-hand axis. For clarity, only the normal and log-normal data is shown. The curves are offset by 0.05 with the actual zero shown by the horizontal line.

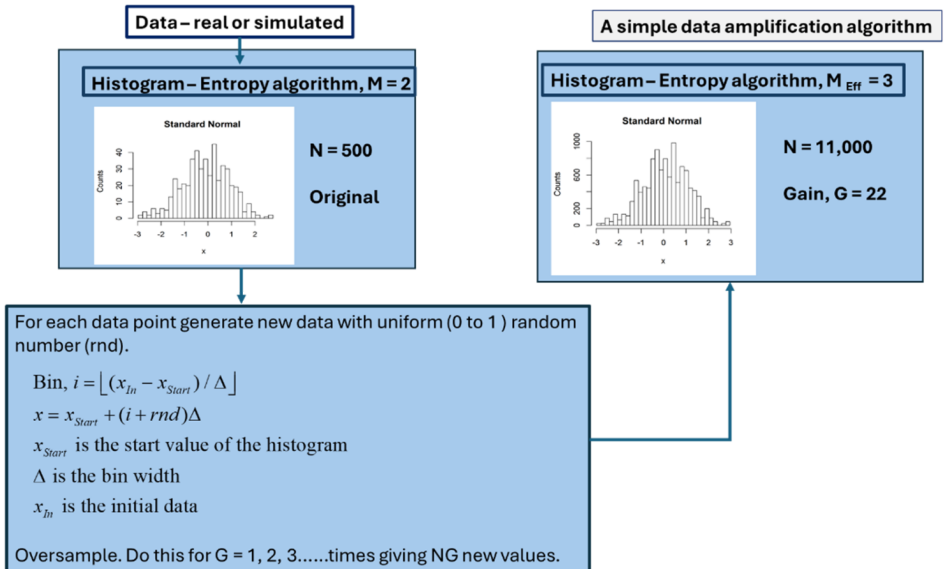

**Figure 2** A simple data amplification algorithm to illustrate the underlying ideas. See text for detail.

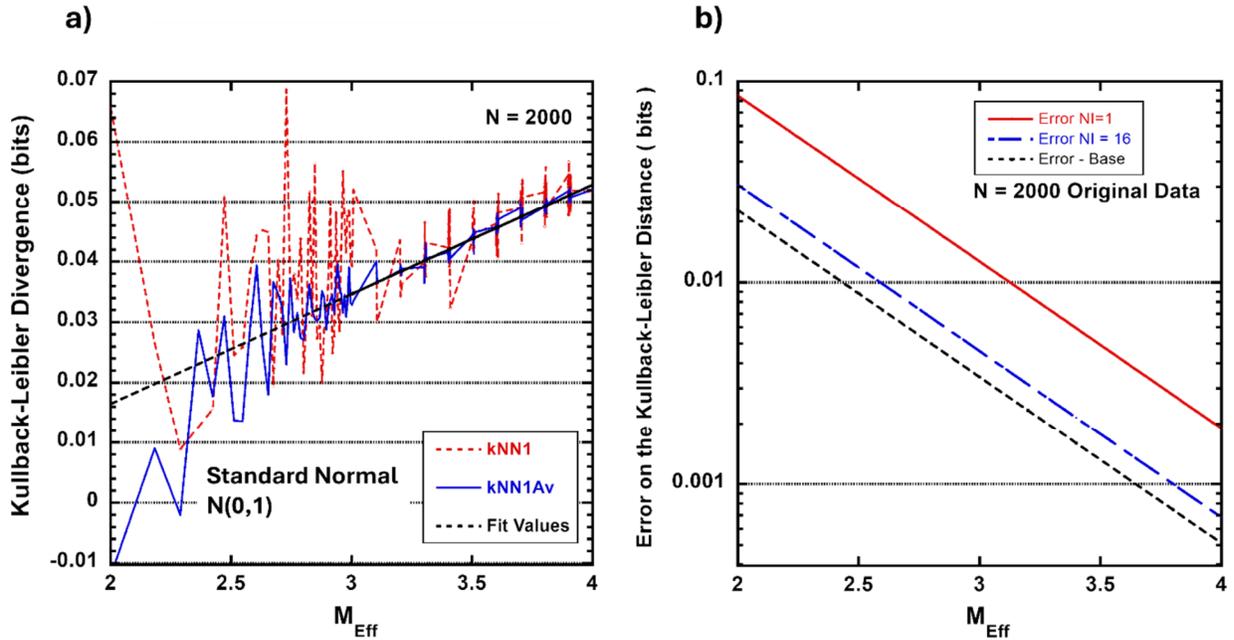

**Figure 3** a) Kullback-Leibler divergence between the amplified data (NG) and an independently generated sample with the same number of events (NG), as a function of $M_{eff}$ for a standard normal distribution. Both a single and averaged estimate are shown. b) Error in bits of the Kullback-Leibler divergence versus $M_{eff}$. The error is for one iteration ($N_I = 1$) and 16 iterations ($N_I = 16$). The base error is the floor below which further iteration gives no improvement, Eq. (2.21). Both figures are for training data, $N = 2000$. See text for more detail.

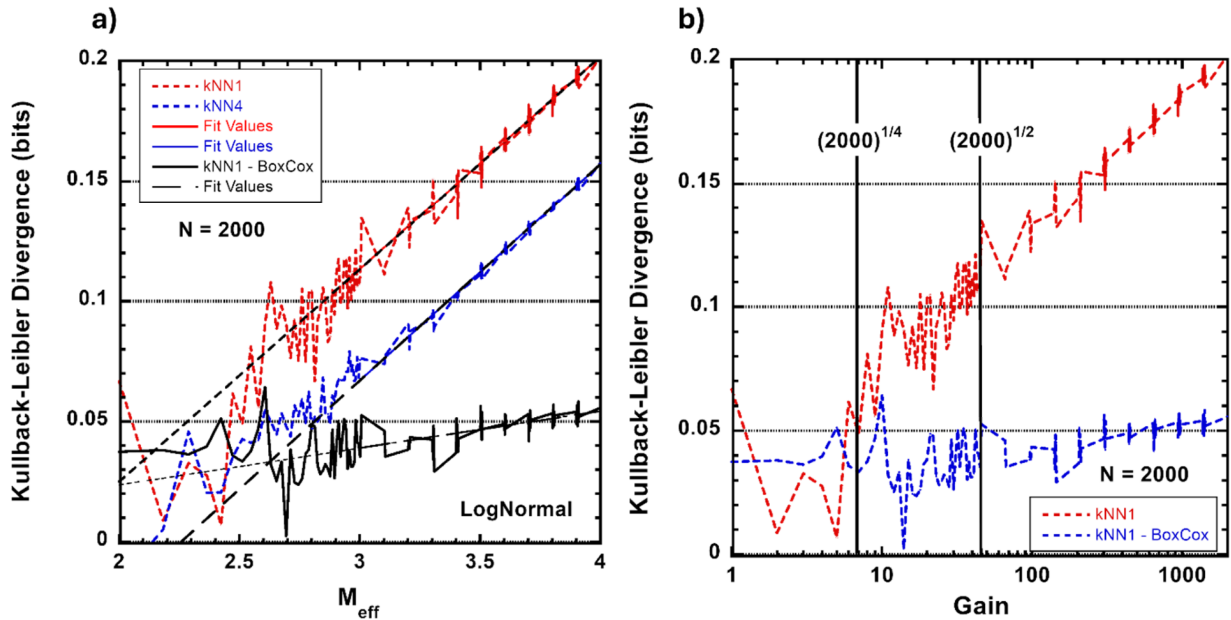

**Figure 4** a) Kullback-Leibler divergence between the amplified data and an independently generated sample with the same number of events as a function of $M_{eff}$. For a standard log-normal distribution. The $kNN = 1$ and $kNN = 4$ values are shown. The $kNN = 1$ values for a distribution amplified after a Box-Cox transformation are also shown. b) Same results but plotted as a function of the gain for $kNN = 1$ only. Lines drawn at $\sqrt[4]{2000}$ and $\sqrt[2]{2000}$ indicate values of $M_{eff}$ corresponding to 2.5 and 3.0 respectively. See text for a full discussion.

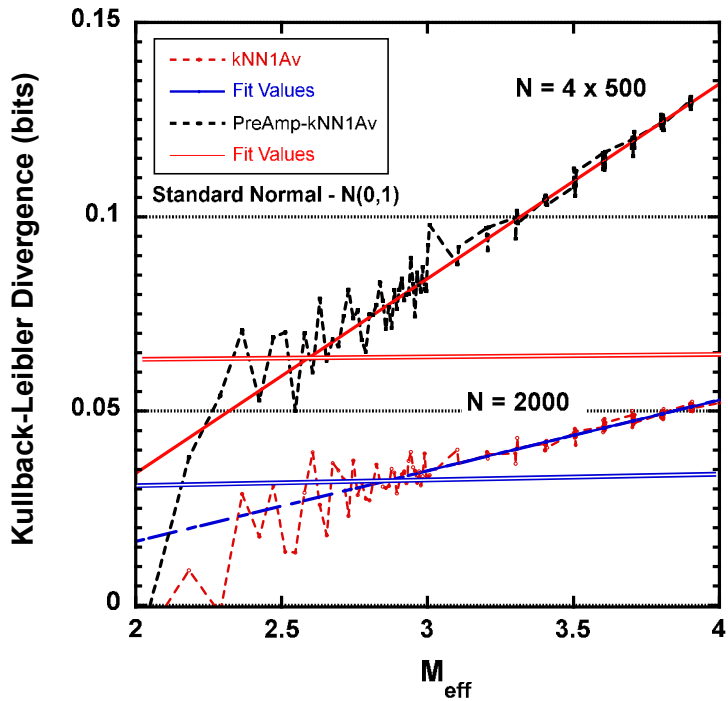

**Figure 5** Kullback-Leibler divergence – generated compared to simulated - for standard normal distributions with initial $N = 2000$ data. The upper curve has training data with 500 entries which is then pre-amplified by four to 2000. The lower curve is for training data of 2000 entries. Both initial samples with 2000 entries are then amplified to show the effect of using pre-amplified data. The double lines indicate the region where the divergence is caused by a poor modelling of the pdf tail. This figure shows that the pre-amplified data has a memory of the starting resolution set by the initial data.

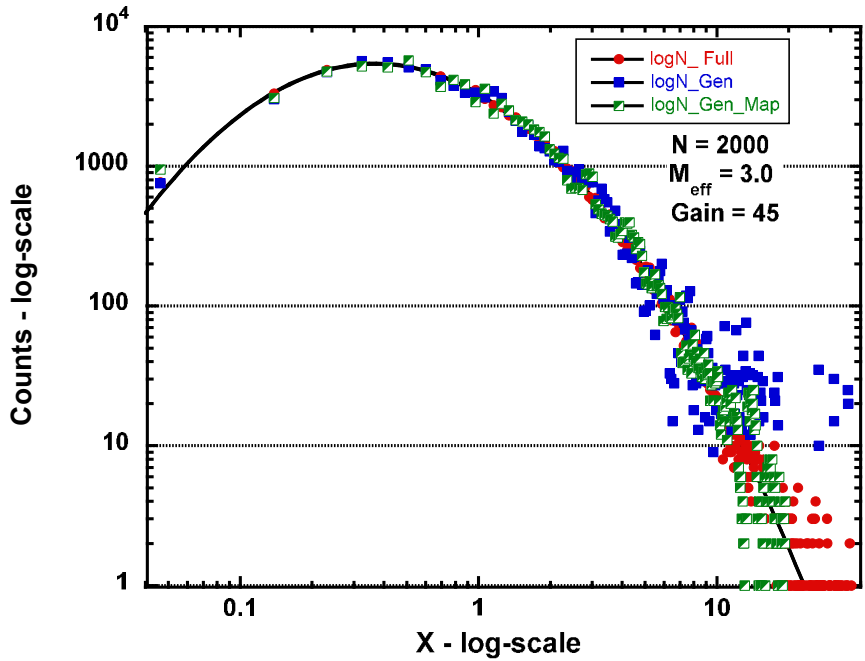

**Figure 6** Comparison of the simulated log-normal distribution for independently simulated (Full), generated from 2000 initial events (Gen) and 2000 initial events to which a Box-Cox transformation was applied (Gen_Map). There are 90,000 "Full" and "Gen" events. This corresponds to $M_{eff} = 3$ and gain, $G = 45$. The fit line is for a standard log-normal and only the overall amplitude parameter was allowed to change. See text for further detail.

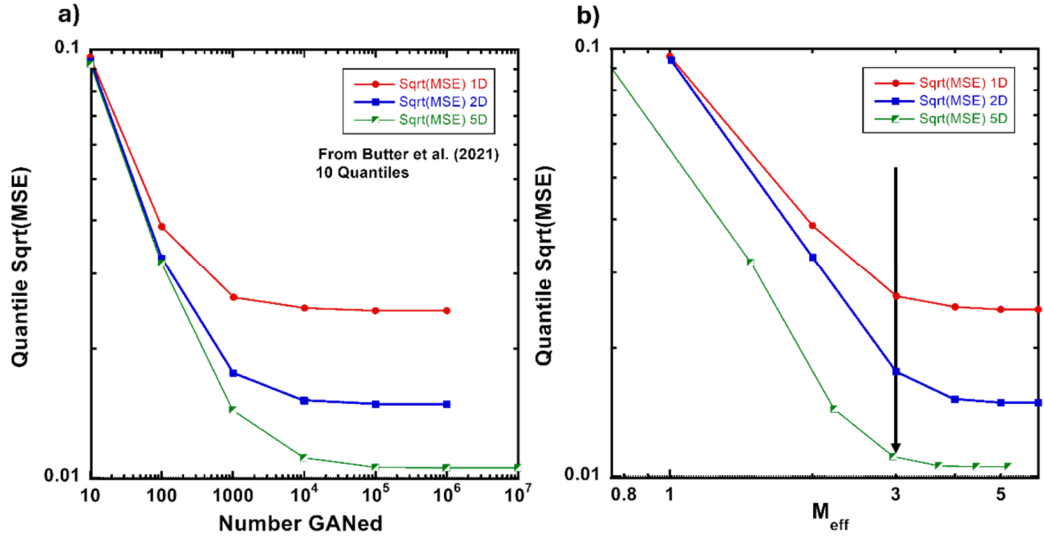

**Figure 7** a) Error for three toy models learnt using a Generative Adversarial Network (GAN) as a function of the number of events produced by a GAN from Butter et al. [9]; 1D closed circle is Fig. 2 left, for a camel back function, 2D closed square is Fig. 4 left for a 2D Gaussian ring (radius), 5D square half filled is Fig. 6 left top (radius). $N = 100$ for the 1D and 2D models. $N = 500$ for the 5D model. 10 quantiles were used for all cases. b) Same data but plotted as a function of $M_{eff} = 2\log(\text{Generated Events})/\log(\text{Trained Events})$. The error reaches a plateau at $M_{eff} = 3$ indicated by the arrow.

**Appendix A**

This section derives the key result on the Shannon entropy of a histogram.

**Definitions** The Shannon entropy of a histogram with $N_{Bin}$ bins is , [A1],

$$H \equiv -\sum_{i=1}^{i=N_{Bin}} p_i \log p_i \quad \text{where the probability for the } i^{th} \text{ bin is } p_i$$

For the $i^{th}$ bin the expected content, $\mu_i = p(x_i)\Delta_i N$ where,

$N$ is the total number of entries in the histogram,
$p(x_i)$ is the probability density function (pdf) at the bin centre, $x_i$, for the $i^{th}$ bin.
$\Delta_i$ is the bin width for the $i^{th}$ bin. Take the bin width, $\Delta$, to be fixed for all bins.
$p_i = p(x_i)\Delta$ is estimated from data using $n_i/N$ where $n_i$ is the measured number of entries in the $i^{th}$ bin. The estimate of $\mu_i$ is $n_i$ .

The estimate of the histogram entropy, is $\quad H_B = -\sum_{i=1}^{i=N_{Bin}} \frac{n_i}{N} \log \frac{n_i}{N}$

**Lemma 1.1** The Shannon entropy of a histogram with a uniform pdf can be written as $H = \frac{1}{M}\log N$ , with $M \geq 1$ and $N$ the total number of entries in the histogram.

**Proof:** For a uniform distribution, set the number of bins, $N_{Bin}$ to $N^{1/M}$. Thus the average number of entries per bin, $\mu_H = N / N^{1/M} = N^{1-1/M}$.

Thus $H = -\sum_{i=1}^{i=N_{Bin}} p_i \log p_i = -N^{1/M} \times \frac{\mu_H}{N} \log \frac{\mu_H}{N} = \log \frac{N}{\mu_H} = \frac{1}{M}\log N$

The minimum number of entries per bin is one which corresponds to $M = 1$ . □

**Lemma 1.2** The continuous, $r_q$ , and discrete form, $R_q$ , of the Rényi entropy are related by $r_q = R_q + \log(\Delta)$. For $q = 1$ , $h = H + \log(\Delta)$ . The differential entropy, $h$ , is defined in the main text.

**Proof:** The Rényi entropy of order $q$ are defined as follows, [A2]. As with the Shannon entropy, capital and lower case will be used for the discrete and continuous versions respectively.

$$R_q \equiv \tfrac{1}{1-q}\ln\left(\sum_{i=1}^{i=N_{Bin}}(p_i)^q\right) \quad \text{and} \quad r_q \equiv \tfrac{1}{1-q}\ln\left(\int_S (p(x))^q dx\right)$$

$r_q = \tfrac{1}{1-q}\ln\left(\int_S (\tfrac{p_i}{\Delta})^q dx\right)$ which one can convert to a Riemann sum

$r_q = \tfrac{1}{1-q}\ln\left(\sum_{i=1}^{i=N_{Bin}}\tfrac{(p_i)^q}{\Delta^{q-1}}\right) = \tfrac{1-q}{1-q}\log(\Delta) + \tfrac{1}{1-q}\ln\left(\sum_{i=1}^{i=N_{Bin}}(p_i)^q\right) = R_q + \log(\Delta)$

The Rényi entropy is related to the Shannon entropy; $R_1 = H$ and $r_1 = h$ from which, $h = H + \log(\Delta)$, which is a well known relation, [A1]. □

There are special cases of the Rényi entropy; $q = 0, 1, 2$ which are also called the max-entropy or Hartley entropy, Shannon entropy, and quadratic or collision entropy, respectively. There is an important ordering, [A3], of the commonly named *spectra of Rényi information*, $R_\beta(X) \leq R_\alpha(X)$ for $\alpha \leq \beta$ with equality only for a uniform distribution. This is also true for the continuous or differential Rényi entropy, $r_q$.

**Theorem 1** The Shannon entropy of a histogram with any pdf can be written as $H = \frac{1}{M} \log N$, with $M \geq 1$ and $N$ the total number of entries in the histogram.

**Proof:** This statement is true for a uniform pdf as shown in Lemma 1.1. In general, consider the probability density function for the number of entries per bin, $q(\mu)$. This is the distribution associated with $\mu_i$ described above. Each has a Poisson distribution, so $\mu$ is a Poisson mixture. Poisson mixtures have a special property, which is shown using the Method of Moments, [A4].

$$E(\mu) = \sum_{i=1}^{i=N_{Bin}} p_i \mu_i \equiv \mu'_H$$ which is the weighted entries/bin. $\mu'_H = \mu_H$ only for a uniform distribution.

$Var(\mu) = E(\mu) + Var(\mu_i)$   Note that $\frac{Var(\mu)}{E(\mu)} > 1$ which is an "over-dispersed" distribution.

$$\mu'_H = \sum_{i=1}^{i=N_{Bin}} p_i \mu_i = N \sum_{i=1}^{i=N_{Bin}} p_i^2$$ which can written in terms of the Rényi Entropy for $q = 2$, as follows,

$\mu'_H = N \exp(-R_2)$. Using Lemma 1.2, $\mu'_H = N\Delta \exp(-r_2)$

Using Lemma 1.2 again, one can write $\Delta = \exp(r_1 - R_1) = \exp(h - H)$ from which

$$\mu'_H = N\Delta \exp(-r_2) = N \exp(-H) \exp(h - r_2) = N \exp(-H) F$$

where $F \equiv \exp(h - r_2)$ is the only part that depends on the underlying pdf. This formula for $\mu'_H$ is for all pdf's including a uniform pdf. Moreover, $R_1 \geq R_2$ and $r_1 \geq r_2$ with equality only for a uniform pdf, see above.

For a uniform pdf, $h = r_2$, so $F = 1$ and $\mu'_H = N \exp(-H)$. But for the uniform distribution, $\mu'_H = \mu_H = N \cdot N^{-1/M}$, Lemma 1.1. Thus, $N \exp(-H) = NN^{-1/M}$ or $H = \frac{1}{M} \log N$ for all distributions. □

**Corollary 1.1** The fixed bin width for a histogram with $N$ entries and differential entropy, $h$, is $\Delta = \exp(h - H) = \frac{\exp(h)}{N^{1/M}}$.

**Proof:** This follows directly from Theorem 1 and Lemma 1.2. See the main text for a discussion on the choice of $M$. □

**Comment**

i) Song [A5] studied the spectra of Rényi information for a large range of distributions. This allows one to calculate the theoretical value of F. Song noted an important parameter, which is called Song's measure,

$$S \equiv \lim_{q \to 1} \frac{\partial r}{\partial q} = -\frac{1}{2} Var(\log(p(x))).$$

This function also appears in the variance analysis of the kNN estimate of $h$, Eq. (2.19). Song realised this is a measure of the intrinsic shape of the pdf, is related to the kurtosis, and provides a partially ordering of pdfs with respect to the tails of their distributions. Clearly, F and S are related and pdfs with a larger kurtosis also give a larger F. Some values of F and S are given in Table A1 below. For these values $F \sim 1 + (1/3)S$. The log-normal distribution has a kurtosis that grows exponentially when $\sigma > 1$ as does F.

ii) F can be estimated from data in two ways. $F_\mu = N^{1/M} \exp(-R_2)$ or $F_E = \exp(H - R_2)$ where $H$ and $R_2$ are estimated using $p_i \simeq n_i/N$. These are sensitive to the pdf being approximated correctly and are best evaluated for $2 \leq M \leq 3$.

Table A1  Shape parameters S and F

| Distribution | $r_1 - r_2$ | Song, S | $F = \exp(r_1 - r_2)$ | Comment |
|---|---|---|---|---|
| Uniform | 0 | 0 | 1 | No scale or location dependence |
| Normal | $0.5(1 - \log(2))$ | 0.5 | 1.1658 | No scale or location dependence. |
| Exponential | $1 - \log(2)$ | 1.0 | 1.359 | No scale dependence. |
| Moyal | 0.216 | 0.734 | 1.2414 | Numerical calc. |
| Log Normal | $0.5(1 - \log(2)) + (\sigma^2/4))$ | $0.5 + \sigma^2$ | 1.4969 $\sigma = 1$ | Scale dependent. |